\documentclass[letterpaper]{article} 
\usepackage[preprint]{aaai2027}  
\usepackage[hyphens]{url}  
\usepackage{graphicx} 
\usepackage{natbib}  
\usepackage{caption} 
\usepackage{amsmath}
\usepackage{amssymb}
\usepackage{multirow}
\usepackage{booktabs}

\title{Adversarial Closed-Loop Curriculum for Evolving Role-Playing Agents}
\author{
    Zheng Zhang\textsuperscript{\rm 1},
    Liu Liu\textsuperscript{\rm 2},
    Qi Chai\textsuperscript{\rm 1},
    Deheng Ye\textsuperscript{\rm 2},
    Peilin Zhao\textsuperscript{\rm 3},
    Mao Zheng\textsuperscript{\rm 2},
    Hao Wang\textsuperscript{\rm 1}
}
\affiliations{
    \textsuperscript{\rm 1}The Hong Kong University of Science and Technology (Guangzhou),\\
    \textsuperscript{\rm 2}Tencent,\\
    \textsuperscript{\rm 3}Shanghai Jiao Tong University
}

\begin{document}

\maketitle

\begin{abstract}
Role-playing agents based on large language models have been widely applied in areas such as personalized assistance and social simulation. 
Recent RL methods typically train on a fixed scenario pool collected before learning begins. 
This creates a distributional bottleneck: as the agent improves, the scenarios where it performs poorly also change, while the training distribution remains static. 
Therefore, we propose AdvRole, an adversarial context rewriting framework that turns role-playing RL into a closed-loop curriculum. 
AdvRole alternates between an Actor that learns to role-play and a Rewriter that edits character profiles and dialogue contexts into actor-specific hard scenarios. 
The Rewriter is trained with a performance-gap reward, which favors rewrites that reduce the current Actor's score relative to the original scenario. 
As a result, the scenario pool evolves with the Actor and continuously targets under-mastered regions of the character-context space. 
Experiments on three role-playing benchmarks covering English and Chinese, as well as a new multilingual benchmark we release, show that AdvRole consistently outperforms baselines.


\end{abstract}

\section{Introduction}

\begin{figure}[t]
\centering
\includegraphics[width=0.85\columnwidth]{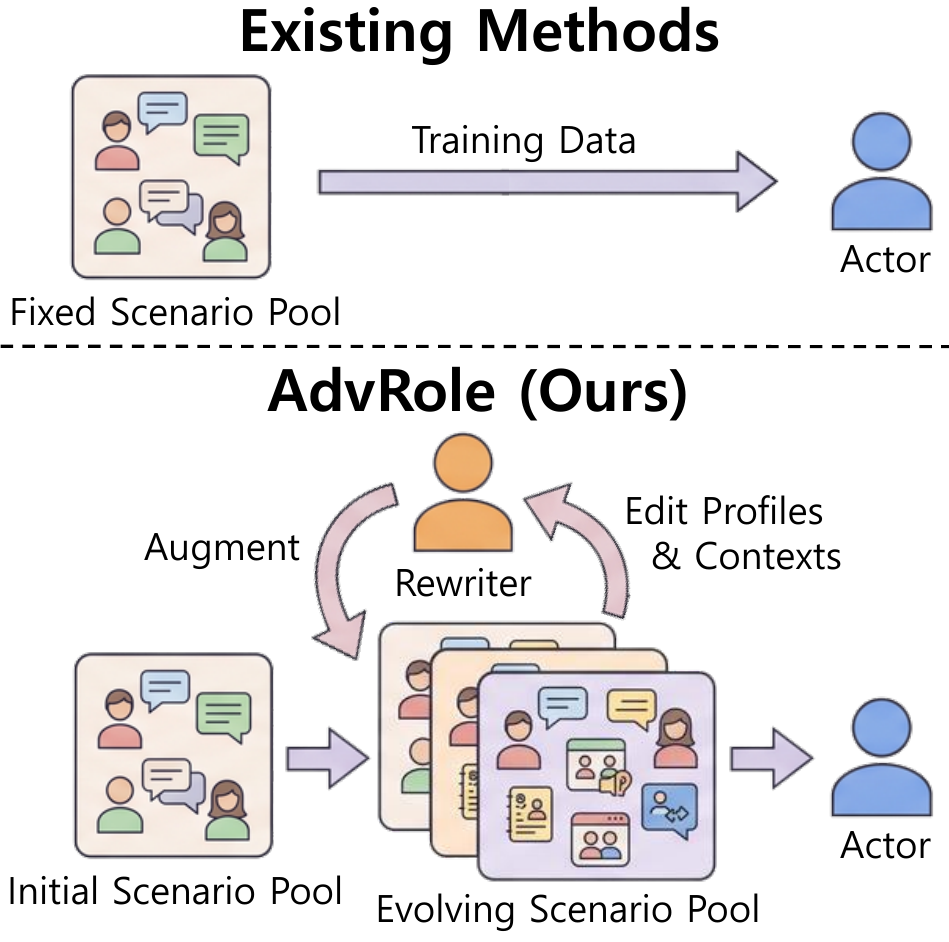}
\vspace{-0.1cm}
\caption{\textbf{Comparison between existing methods and AdvRole.} Existing methods train the Actor on a fixed scenario pool collected before training. In contrast, AdvRole introduces a Rewriter that continually edits character profiles and dialogue contexts to expose Actor failures, evolving the scenario pool throughout training.}
\label{fig:concept}
\vspace{-0.2cm}
\end{figure}

Large language models (LLMs) have demonstrated remarkable capabilities in open-ended generation, ranging from creative writing~\cite{chung2025modifying, wei-etal-2025-igniting, rashkin-etal-2025-help, saakyan2026death} to emotional support~\cite{chen-etal-2023-soulchat, 10.1145/3706598.3713453, yuan-etal-2025-reflectdiffu, wang-etal-2025-flexible} and communication game~\cite{lan-etal-2024-llm, 10888525, 10.1145/3746027.3755752}. Among these tasks, role-playing has attracted increasing attention~\cite{chen2024roleplaysurvey, wang2024coser, verirole_iclr26, r4_iclr26}, with applications in interactive entertainment, social simulation, and personalized assistants.

In role-playing, an agent is asked to act as a designated character and produce dialogue that is faithful to the character's profile and consistent with the ongoing conversation. Early approaches build role-playing agents through prompt engineering~\cite{wang-etal-2024-rolellm, wang2024incharacter, han-etal-2024-psydial} or supervised fine-tuning on curated dialogue corpora~\cite{shao-etal-2023-character, lu-etal-2024-superpositions, zhou-etal-2024-characterglm, wang2024coser}. More recent work turns to reinforcement learning (RL), starting with PPO or DPO on human-annotated preference data~\cite{apcdpo_neurips24, persllm_emnlp24, pcl_acl25, multiturnrl_neurips25}, and later moving to GRPO~\cite{DBLP:journals/corr/abs-2402-03300} using automatically computed rewards that remove the need for human annotation~\cite{cogdual_emnlp25, wang2025raidenr1improvingroleawarenessllms, verirole_iclr26, r4_iclr26}.
Despite these advances, most RL-based methods remain reward-centric: they improve how responses are scored or optimized, but train on a fixed scenario pool collected before learning begins.

This static training distribution creates a mismatch between optimization and data.
A role-playing scenario is the joint product of a character profile and a dialogue context, covering factors such as personality, background, speaking style, scene, partner, and dialogue history.
More importantly, scenario difficulty is actor-dependent.
As the Actor improves, the scenarios where it still struggles also change.
Common profiles and contexts may become uninformative, while useful training data shifts toward rarer traits, subtle conflicts, and unfamiliar dialogue states.
Thus, the key question is not only how to reward a response, but also where to train the Actor next.
A fixed pool cannot follow this moving failure boundary.

To this end, we propose AdvRole, an adversarial closed-loop curriculum for evolving role-playing agents. As shown in Figure~\ref{fig:concept}, 
AdvRole alternates between two models: an \textbf{Actor} that learns to perform role-play and a \textbf{Rewriter} that edits character profiles and dialogue contexts into actor-specific hard scenarios.
The Actor is trained with a standard role-playing reward model.
The Rewriter is optimized with a performance-gap reward: a rewritten scenario is considered useful when it causes the current Actor to receive a lower role-playing score than on the original scenario.
In this way, the Rewriter acts as a diagnostic adversary, searching for plausible scenario variations that reveal where the Actor currently fails.
The resulting scenario pool evolves with the Actor, continuously pushing training toward under-mastered regions of the character-context space.


We evaluate AdvRole on three role-playing benchmarks covering English and Chinese, where it outperforms baselines. To enable broader evaluation, we also release a new multilingual role-playing benchmark, and further demonstrate the effectiveness of AdvRole on it.

Our contributions can be summarized as follows:
\begin{itemize}
\item We propose AdvRole, an adversarial RL framework that alternately trains an Actor and a Rewriter, continually improving the Actor's role-playing ability.
\item We introduce a reward based on the Actor's performance gap that drives the Rewriter to generate more challenging character profiles and contexts for the Actor.
\item We demonstrate the effectiveness of our framework on three existing benchmarks, and release a new multilingual role-playing benchmark to support broader evaluation.
\end{itemize}

\section{Related Work}

\subsection{Role-Playing Agent}
Role-playing agents are commonly built by prompting LLMs with character profiles, personality traits, and few-shot dialogues~\cite{wang-etal-2024-rolellm, wang2024incharacter, han-etal-2024-psydial}, or by fine-tuning them on role-playing corpora collected from novels, scripts, and synthetic conversations~\cite{shao-etal-2023-character, lu-etal-2024-superpositions, zhou-etal-2024-characterglm, wang2024coser}. 
More recent work improves role-playing agents with reinforcement learning, using PPO or DPO with human preference data~\cite{ouyang2022training, rafailov2023direct, apcdpo_neurips24, persllm_emnlp24, pcl_acl25, multiturnrl_neurips25}, or GRPO with automatically computed rewards~\cite{DBLP:journals/corr/abs-2402-03300, cogdual_emnlp25, wang2025raidenr1improvingroleawarenessllms, verirole_iclr26, r4_iclr26}. 
These methods differ in how they elicit, fine-tune, or reward in-character behavior, but they typically optimize the agent on a fixed scenario pool collected before training. 
AdvRole instead evolves the training distribution itself by adversarially rewriting character profiles and dialogue contexts into actor-specific hard scenarios.

\subsection{Label-Free Reinforcement Learning}
Label-free reinforcement learning (RL) aims to improve LLMs without relying on ground-truth answers or human-annotated preferences. Existing methods derive reward signals from the model's own behavior, such as sequence-level confidence~\cite{li2025confidence, prabhudesai2025confidencerl}, consistency over multiple reasoning paths~\cite{zuo2025ttrl, zhang2025consistentpaths, prasad2024scpo}, or output entropy~\cite{agarwal2025entropy, cheng2025entropy}. Others train on synthetic tasks created by the model itself, using execution-based or self-derived signals as rewards~\cite{zhao2025absolutezero, zhou2025selfchallenging}. These methods are mostly studied in verifiable domains such as math and code, where automatic verification is straightforward. Role-playing is a natural fit for label-free RL, since there is no single ground-truth response to align against and human preference labels are expensive to obtain at scale.

\begin{figure*}[t]
    \centering
    \includegraphics[width=\linewidth]{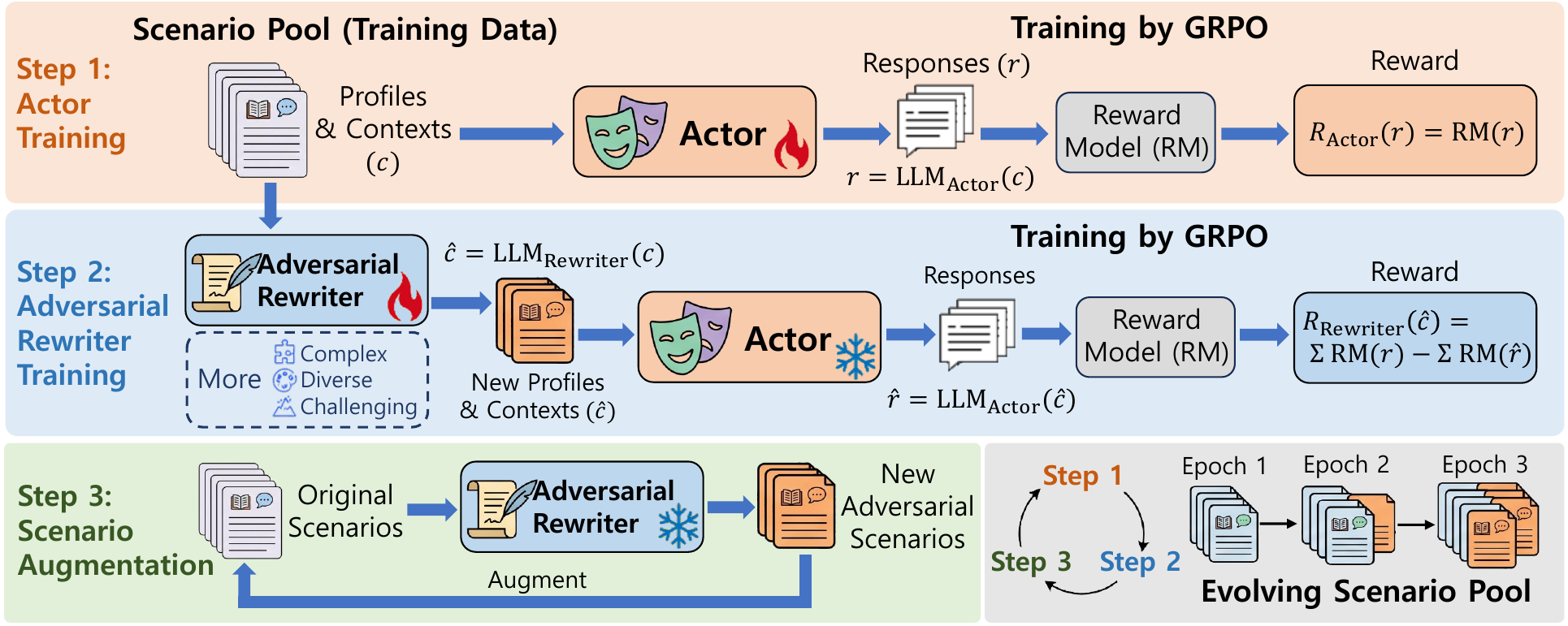}

\caption{\textbf{The training framework of AdvRole.} In each epoch, the Actor is first trained by GRPO on the current scenario pool with rewards from a reward model, and the Rewriter is then trained to edit original scenarios to expose the Actor's failures, with its reward defined by the Actor's performance gap between the original and rewritten scenarios. The updated Rewriter further augments the pool for the next epoch, yielding an evolving scenario pool.}

    \label{fig:framework}
    \vspace{-0.2cm}
\end{figure*}

\subsection{Self-Play in LLMs}
Self-play, where two roles co-evolve through their interaction, has recently emerged as a promising paradigm for improving LLMs without external supervision or human-labeled data. In verifiable domains, a generator and a verifier are paired so that the verifier's automatic judgment provides reward for the generator, as in coder-tester frameworks for code generation~\cite{lin2025solveverify, jiang2025verse} and proposer-solver frameworks for math reasoning~\cite{zhao2025absolutezero, rzero_iclr26}. Our framework follows this self-play paradigm but targets role-playing, where outputs cannot be checked by a deterministic verifier.

\section{Method}

\subsection{Overview}

AdvRole trains two models alternately: an \textbf{Actor} that performs role-playing, and a \textbf{Rewriter} that edits the original character profile and dialogue context to expose the Actor's failures. Both models are optimized with GRPO~\cite{DBLP:journals/corr/abs-2402-03300}, and the scenario pool for training keeps evolving as the Rewriter improves.

As shown in Figure~\ref{fig:framework}, each epoch of AdvRole consists of three steps: (1) \textbf{Actor Training}: the Actor is updated on the current scenario pool using a reward model that scores role-playing quality. (2) \textbf{Adversarial Rewriter Training}: the Rewriter is updated with a reward defined by the Actor's performance gap between the original and rewritten scenarios, encouraging it to expose the weaknesses of the current Actor. (3) \textbf{Scenario Augmentation}: the updated Rewriter rewrites the original scenarios into variants that reveal failures, which are merged into the pool to train the Actor in the next epoch.

\subsection{Actor Training}
\label{sec:actor_training}

We formalize a role-playing scenario as a tuple $c = (p, d)$, where $p$ is the character profile and $d$ is the dialogue context that the agent needs to respond to. Given the current scenario pool $\mathcal{S}$, the Actor $\pi_\phi$ generates a response conditioned on $c$:
\begin{equation}
r \sim \pi_\phi(\cdot \mid c), \quad c \in \mathcal{S}.
\end{equation}

To assess the role-playing quality of $r$, we follow~\citet{cpo_emnlp25} and adopt a strong open-source LLM (Qwen2.5-72B-Instruct\footnote{\url{https://huggingface.co/Qwen/Qwen2.5-72B-Instruct}}) as the reward model (RM), which scores $r$ along several role-playing dimensions given the scenario $c$ (the full scoring prompt is provided in the supplementary material):
\begin{equation}
\label{eq:actor_reward}
R_{\text{Actor}}(r) = \mathrm{RM}(r \mid c).
\end{equation}

To avoid overly long responses, we multiply the Actor reward by a decay coefficient based on the response length $|r|$ (in words for English and in characters for Chinese). The coefficient is $1$ below a lower threshold $\tau_1$, decreases linearly to $0$ at an upper threshold $\tau_2$, and remains $0$ beyond $\tau_2$, with $(\tau_1, \tau_2) = (50, 100)$ for English and $(60, 120)$ for Chinese.

We optimize the Actor with GRPO~\cite{DBLP:journals/corr/abs-2402-03300}. For a policy $\pi_\theta$ trained on inputs $x$ with reward $R$, GRPO samples a group of $G$ outputs $\{o^{(i)}\}_{i=1}^G \sim \pi_\theta(\cdot \mid x)$ and maximizes:
\begin{equation}
\label{eq:grpo}
\begin{aligned}
\mathcal{J}(\theta) &= \mathbb{E}_{x} \left[ \frac{1}{G} \sum_{i=1}^G \mathcal{L}_i^{\theta}(A^{(i)}) - \beta D_{KL}(\pi_\theta \,||\, \pi_{\text{ref}}) \right], \\
A^{(i)} &= \frac{R(o^{(i)}) - \mu}{\sigma},
\end{aligned}
\end{equation}
where $\mathcal{L}_i^{\theta}(A^{(i)})$ is the clipped surrogate objective parameterized by the normalized advantage $A^{(i)}$, and $\mu$, $\sigma$ are the mean and standard deviation of the group rewards. For the Actor, we instantiate $\pi_\theta$ as $\pi_\phi$, $x$ as the scenario $c$, and $R$ as $R_{\text{Actor}}$ in Equation~(\ref{eq:actor_reward}).

\begin{table*}[t]
\centering
\small
\begin{tabular}{lcccc}
\toprule
\multirow{2}{*}{\textbf{Model}} & \textbf{Conversational} & \textbf{Character} & \textbf{Role-playing} & \multirow{2}{*}{\textbf{Overall} ($\uparrow$)} \\
 & \textbf{Ability} ($\uparrow$) & \textbf{Consistency} ($\uparrow$) & \textbf{Attractiveness} ($\uparrow$) & \\
\midrule
\multicolumn{5}{l}{\textit{Large-Scale LLMs}} \\

\quad Kimi-K2.6 (non-thinking) & 4.10 & 3.21 & 3.43 & 3.58 \\ 
\quad Kimi-K2.6 (thinking) & 4.06 & 3.35 & 3.58 & 3.66 \\ 
\quad Claude-3.7-Sonnet & 3.90 & 3.06 & 3.25 & 3.41 \\ 
\quad GPT-4o & 3.63 & 2.90 & 3.06 & 3.20 \\ 
\quad Qwen2.5-72B-Instruct & 3.80 & 3.05 & 3.28 & 3.37 \\

\midrule
\multicolumn{5}{l}{\textit{Qwen2.5-7B-Instruct}} \\
\quad Vanilla                       & 3.71 & 2.91 & 3.13 & 3.25 \\
\quad CPO~\cite{cpo_emnlp25}       & 3.75 & 3.03 & 3.30 & 3.36 \\
\quad R4~\cite{r4_iclr26}          & 3.79 & 2.79 & 3.05 & 3.21 \\
\quad VeriRole~\cite{verirole_iclr26} & 3.50 & \textbf{3.16} & 3.24 & 3.30 \\
\quad \textbf{AdvRole (Ours)}          & \textbf{3.83} & 3.10 & \textbf{3.36} & \textbf{3.43} \\
\midrule
\multicolumn{5}{l}{\textit{Qwen2.5-14B-Instruct}} \\
\quad Vanilla                       & 3.72 & 2.95 & 3.17 & 3.28 \\
\quad CPO~\cite{cpo_emnlp25}       & 3.81 & 3.04 & 3.29 & 3.38 \\
\quad R4~\cite{r4_iclr26}          & 3.86 & 3.00 & 3.22 & 3.36 \\
\quad VeriRole~\cite{verirole_iclr26} & 3.56 & 3.17 & 3.23 & 3.32 \\
\quad \textbf{AdvRole (Ours)}          & \textbf{3.89} & \textbf{3.18} & \textbf{3.40} & \textbf{3.49} \\
\bottomrule
\end{tabular}
\caption{\textbf{Performance on the CharacterEval benchmark.} Each dimension score is the arithmetic mean of its underlying metrics between $1$ and $5$, and ``Overall'' is the mean of the three dimensions. All baselines re-implemented under identical data splits.}
\label{tab:charactereval}
\vspace{-0.2cm}
\end{table*}

\subsection{Adversarial Rewriter Training}
\label{sec:rewriter_training}

While the Actor is trained on $\mathcal{S}$, the pool itself has limited coverage and diversity. As the Actor gradually fits this pool, further training on the same data brings limited gain. To break this bottleneck, we introduce a Rewriter $\pi_\psi$ that edits an original scenario $c = (p, d)$ into a rewritten one $\hat{c} = (\hat{p}, \hat{d})$:
\begin{equation}
\hat{c} \sim \pi_\psi(\cdot \mid c).
\end{equation}

The Rewriter is prompted to make $\hat{c}$ more complex and more diverse than $c$, so that it covers scenarios underrepresented in the original pool $\mathcal{S}$ and exposes the Actor's weaknesses.

To train the Rewriter to expose the weaknesses of the current Actor, we define its reward as the gap between the Actor's performance on the original and rewritten scenarios. Specifically, we freeze the Actor and sample $N$ responses ($N{=}3$ in our experiments) for each scenario:
\begin{equation}
\label{eq:sample_n}
\{r^{(i)}\}_{i=1}^N \sim \pi_\phi(\cdot \mid c), \quad \{\hat{r}^{(i)}\}_{i=1}^N \sim \pi_\phi(\cdot \mid \hat{c}).
\end{equation}

The Rewriter's reward is then computed as:
\begin{equation}
\label{eq:rewriter_reward}
\begin{split}
R_{\text{Rewriter}}(\hat{c}) = {} & \frac{1}{N}\sum_{i=1}^N \mathrm{RM}(r^{(i)} \mid c) \\
& - \frac{1}{N}\sum_{i=1}^N \mathrm{RM}(\hat{r}^{(i)} \mid \hat{c}).
\end{split}
\end{equation}

Averaging over multiple responses helps reduce the variance from stochastic decoding and reward-model scoring, yielding a more stable estimate of the difficulty gap. A larger $R_{\text{Rewriter}}(\hat{c})$ indicates that the rewritten scenario reveals the weakness for the current Actor, encouraging the Rewriter to discover character profiles and contexts that the Actor has not yet mastered.

We also constrain the length of rewritten scenarios. If $|\hat{c}|$ exceeds $150\%$ of the original length $|c|$, we set $R_{\text{Rewriter}}(\hat{c})$ to $-5$, which discourages the Rewriter from padding scenarios with irrelevant content.

We then update the Rewriter by maximizing the GRPO objective in Equation~(\ref{eq:grpo}), with $\pi_\theta$ instantiated as $\pi_\psi$, $x$ as the original scenario $c$, and $R$ as $R_{\text{Rewriter}}$ in Equation~(\ref{eq:rewriter_reward}).

\subsection{Scenario Augmentation}
\label{sec:scenario_augmentation}

After updating the Rewriter, we use it to expand the scenario pool. For each original scenario $c \in \mathcal{S}_0$ in the initial pool, the updated Rewriter produces a rewritten version $\hat{c}$, and all rewritten scenarios are merged into the pool to form the training data for the next epoch:
\begin{equation}
\mathcal{S} \leftarrow \mathcal{S} \cup \{\hat{c} \mid \hat{c} \sim \pi_\psi(\cdot \mid c),\, c \in \mathcal{S}_0\}.
\end{equation}
We always rewrite from the original $\mathcal{S}_0$ rather than from previously rewritten scenarios, which keeps each epoch's augmentation grounded in real character profiles and contexts and avoids drifting too far from the data distribution.

By alternating the three steps across epochs, the scenario pool keeps evolving alongside the Actor, and the Actor is continuously exposed to more diverse scenarios that target its current weaknesses.

\section{Experiments}

\subsection{Implementation Details}
\label{sec:imp_detail}

We separately train and evaluate AdvRole on two role-playing benchmarks of different languages: CoSER~\cite{coser_icml25} for English and CharacterEval~\cite{charactereval_acl24} for Chinese. To further assess generalizability, we additionally take the model trained on CharacterEval and evaluate it in a zero-shot manner on RAIDEN~\cite{wu-etal-2025-raiden}, another Chinese role-playing benchmark.

For CoSER, we follow CogDual~\cite{cogdual_emnlp25} and use 17,762 scenarios as the training set and 200 as the test set. For CharacterEval, we randomly select 2,000 scenarios as the training set and use the remaining 2,564 as the test set. For RAIDEN, we directly evaluate on its test set.

\begin{table*}[t]
\centering
\small
\setlength{\tabcolsep}{4pt}
\renewcommand{\arraystretch}{1.15}
\begin{tabular}{l c c c c c c c}
\toprule
\multirow{2}{*}{\textbf{Model}} &
\textbf{Storyline} & \textbf{Anthropo-} & \textbf{Character} & \textbf{Storyline} &
\textbf{Average} & \textbf{BLEU} & \textbf{ROUGE-L} \\
 & \textbf{Consistency} ($\uparrow$) & \textbf{morphism} ($\uparrow$) & \textbf{Fidelity} ($\uparrow$) & \textbf{Quality} ($\uparrow$) & ($\uparrow$) & ($\uparrow$) & ($\uparrow$) \\
\midrule
\multicolumn{8}{l}{\textit{Large-Scale LLMs}} \\
\quad Gemini-1.5-Pro           & 59.1 & 52.4 & 47.8 & 77.6 & 59.2 & 0.054 & 0.117 \\
\quad DeepSeek-V3              & 56.4 & 47.9 & 44.0 & 76.7 & 56.2 & 0.045 & 0.110 \\
\quad Claude-3.5-Sonnet        & 57.5 & 48.5 & 45.7 & 77.2 & 57.2 & 0.052 & 0.115 \\
\quad GPT-4o                   & 58.9 & 43.1 & 41.6 & 75.4 & 54.8 & 0.059 & 0.121 \\

\midrule
\multicolumn{8}{l}{\textit{Qwen2.5-7B-Instruct}} \\
\quad Vanilla              & 59.9 & 42.0 & 41.5 & 62.3 & 51.4 & 0.007 & 0.155 \\
\quad CoT                  & 55.8 & 37.2 & 36.5 & 61.8 & 47.8 & 0.007 & 0.153 \\
\quad CB-CoT               & 56.9 & 44.9 & 39.1 & 62.5 & 50.8 & 0.008 & 0.155 \\
\quad CogDual-SFT          & 58.4 & 47.0 & 45.0 & 71.7 & 55.5 & 0.015 & 0.161 \\
\quad CogDual-RL           & 59.9 & 46.6 & 47.0 & 74.0 & 56.9 & \textbf{0.016} & \textbf{0.163} \\
\quad \textbf{AdvRole (Ours)} & \textbf{61.5} & \textbf{48.5} & \textbf{49.2} & \textbf{77.2} & \textbf{59.1} & 0.015 & 0.161 \\
\midrule
\multicolumn{8}{l}{\textit{LLaMA3.1-8B-Instruct}} \\
\quad Vanilla              & 48.2 & 36.6 & 27.0 & 63.7 & 43.9 & 0.046 & 0.102 \\
\quad CoT                  & 50.1 & 40.4 & 28.0 & 64.3 & 45.7 & 0.047 & 0.103 \\
\quad CB-CoT               & 52.8 & 41.4 & 27.7 & 65.0 & 46.7 & 0.048 & 0.105 \\
\quad CogDual-SFT          & 56.0 & 46.9 & 43.8 & 75.1 & 55.4 & 0.061 & 0.118 \\
\quad CogDual-RL           & 60.1 & 45.9 & 48.8 & 73.1 & 57.0 & \textbf{0.064} & \textbf{0.121} \\
\quad \textbf{AdvRole (Ours)} & \textbf{61.3} & \textbf{47.2} & \textbf{50.3} & \textbf{75.4} & \textbf{58.6} & 0.060 & 0.119 \\
\bottomrule
\end{tabular}
\caption{\textbf{Performance on the CoSER benchmark.} The first four columns are LLM-judged dimension scores (0--100). “Average” is the arithmetic mean of the four dimension means. BLEU and ROUGE-L are corpus-level n-gram metrics against the reference dialogue.}
\label{tab:coser_main}
\vspace{-0.2cm}
\end{table*}

\subsubsection{Training Scheme}
\label{sec:training_scheme}

We implement our training pipeline based on the \texttt{verl} framework\footnote{\url{https://github.com/verl-project/verl}}~\cite{10.1145/3689031.3696075}. For the GRPO hyperparameter settings, we configure the sample size to $G=8$, the clipping parameter to $\epsilon=0.2$, and the KL divergence coefficient to $\beta=0.04$. The total batch size per step is set to $32$. The Actor and the Rewriter are initialized from the same LLM and fine-tuned as two separate models. For the Rewriter reward in Equation~(\ref{eq:rewriter_reward}), we sample $N=3$ Actor responses per scenario.

The training is conducted with a learning rate of $1\times10^{-6}$ across 16 H20 GPUs, alternating Actor training, Rewriter training, and scenario augmentation for $3$ epochs in total.


\subsubsection{Evaluation Setup}

\paragraph{CoSER.} CoSER evaluates an entire multi-turn dialogue rather than a single response. At test time, an auxiliary LLM simulates both the dialogue partner and the environment, conducting a 30-turn conversation with the role-playing model. An LLM judge then rates the dialogue with a penalty-based protocol along four dimensions: Storyline Consistency, Anthropomorphism, Character Fidelity, and Storyline Quality. Each dimension starts from 100 and is deducted according to detected flaws, so all scores fall in the range $[0, 100]$ and higher is better. Following our baselines~\cite{coser_icml25, cogdual_emnlp25}, we use GPT-4o as both the dialogue partner and the LLM judge.

\paragraph{CharacterEval.} CharacterEval evaluates a single response given a character profile and dialogue context, using BaichuanCharRM\footnote{\url{https://huggingface.co/morecry/BaichuanCharRM}}~\cite{charactereval_acl24} to score 12 metrics on a $1$--$5$ scale. The metrics are grouped into three dimensions: Conversational Ability (Fluency, Coherency, Consistency), Character Consistency (covering both knowledge-related and persona-related aspects), and Role-playing Attractiveness (Human-Likeness, Communication Skills, Expression Diversity, Empathy). We report each dimension as the mean of its metrics and an Overall score as the mean of the three dimensions.

\paragraph{RAIDEN.} RAIDEN evaluates a single response at predefined dialogue anchor points. We follow the evaluation setting of VeriRole~\cite{verirole_iclr26}, which uses an LLM judge (GPT-4o) to assign a binary correctness score on six fine-grained dimensions and reports the score in $[0, 1]$.

All prompts used in training and evaluation are provided in the supplementary material.

\subsection{Main Results}

We evaluate AdvRole on CharacterEval against three recent role-playing RL methods: CPO~\cite{cpo_emnlp25}, R4~\cite{r4_iclr26}, and VeriRole~\cite{verirole_iclr26}. We use Qwen2.5-7B-Instruct\footnote{\url{https://huggingface.co/Qwen/Qwen2.5-7B-Instruct}} and Qwen2.5-14B-Instruct\footnote{\url{https://huggingface.co/Qwen/Qwen2.5-14B-Instruct}} as the base models for fine-tuning.

As shown in Table~\ref{tab:charactereval}, AdvRole achieves the best overall score under both base models, with consistent improvements across most dimensions. The only exception is character consistency under Qwen2.5-7B-Instruct, on which VeriRole is marginally higher. This is likely because VeriRole explicitly optimizes for character consistency through hint-based rewards, whereas AdvRole distributes its training signal over a broader set of role-playing aspects.

More importantly, our fine-tuned Qwen2.5-14B-Instruct surpasses several large-scale LLMs with far more parameters on the overall score. This suggests that the bottleneck of small open-source role-playing models lies less in raw model capacity than in the limited coverage of their training scenarios, and that adversarially expanding the scenario pool is an effective solution.

We further evaluate AdvRole on CoSER against several baselines from CogDual~\cite{cogdual_emnlp25}. As shown in Table~\ref{tab:coser_main}, AdvRole achieves the best average score under both base models, with gains on all four LLM-judged dimensions. Against the strongest baseline CogDual-RL, the gain is the largest on storyline quality and the second largest on character fidelity under both base models. This pattern indicates that the rewritten scenarios mainly help the Actor on character profiles and dialogue contexts that the original pool covers sparsely.

On BLEU and ROUGE-L, AdvRole is slightly below CogDual-RL, as our reward targets role-playing quality rather than n-gram overlap with reference dialogues.

\subsection{Cross-Benchmark Generalization}

\begin{table*}[t]
\centering
\small
\setlength{\tabcolsep}{6pt}
\renewcommand{\arraystretch}{1.15}
\begin{tabular}{l c c c c c c}
\toprule
\textbf{Qwen2.5-7B-Instruct} & \textbf{SBK} ($\uparrow$) & \textbf{CM} ($\uparrow$) & \textbf{SCK} ($\uparrow$) & \textbf{RCB} ($\uparrow$) & \textbf{TA} ($\uparrow$) & \textbf{Average} ($\uparrow$) \\
\midrule
Vanilla                          & 0.69 & 0.66 & 0.61 & 0.48 & 0.42 & 0.57 \\
CPO~\cite{cpo_emnlp25}          & 0.71 & 0.69 & 0.63 & 0.50 & 0.44 & 0.59 \\
R4~\cite{r4_iclr26}             & 0.70 & 0.68 & 0.64 & 0.51 & 0.43 & 0.59 \\
VeriRole~\cite{verirole_iclr26} & \textbf{0.74} & \textbf{0.72} & 0.65 & 0.51 & 0.44 & 0.61 \\
\textbf{AdvRole (Ours)}             & 0.73 & 0.71 & \textbf{0.68} & \textbf{0.58} & \textbf{0.51} & \textbf{0.64} \\

\bottomrule
\end{tabular}
\caption{\textbf{Zero-shot performance on the RAIDEN benchmark.} All models are trained on CharacterEval and directly evaluated on RAIDEN without further tuning. Following VeriRole~\cite{verirole_iclr26}, each dimension is scored by an LLM judge (GPT-4o). The five dimensions are Script-Based Knowledge (SBK), Conversation Memory (CM), Script-Contradictory Knowledge (SCK), Role-Cognition Boundary (RCB), and Topic Advancement (TA).}
\label{tab:raiden}
\vspace{-0.2cm}
\end{table*}

To examine whether the gains brought by AdvRole transfer beyond the training benchmark, we take the model trained on CharacterEval and directly evaluate it on RAIDEN without any further tuning. As shown in Table~\ref{tab:raiden}, AdvRole achieves the best average score. At the dimension level, VeriRole obtains the highest scores on SBK and CM, which are exactly the dimensions its hint-based reward is designed to target through character profiles and dialogue history. In contrast, AdvRole leads on SCK, RCB and TA, which require the model to handle out-of-profile situations and steer the conversation in character.

This matches what each method optimizes for. VeriRole's reward is tied to verifiable hints from profiles and histories, while AdvRole distributes its signal over broader role-playing aspects through rewritten scenarios, leading to stronger behavior beyond profile adherence.

\subsection{Ablation Study}

\begin{table}[t]
\centering
\small
\setlength{\tabcolsep}{3.0pt}
\renewcommand{\arraystretch}{1.1}
\begin{tabular}{lcccc}
\toprule
\textbf{Variant} & \textbf{CA} ($\uparrow$) & \textbf{CC} ($\uparrow$) & \textbf{RA} ($\uparrow$) & \textbf{Overall} ($\uparrow$) \\
\midrule
Profile only         & 3.81 & 3.08 & 3.33 & 3.41 \\
Context only         & 3.79 & 3.07 & 3.31 & 3.39 \\
Profile + Context    & \textbf{3.83} & \textbf{3.10} & \textbf{3.36} & \textbf{3.43} \\
\midrule
w/o Rewriter         & 3.74 & 2.89 & 3.18 & 3.27 \\
Frozen Rewriter      & 3.77 & 3.05 & 3.28 & 3.37 \\
Co-trained Rewriter  & \textbf{3.83} & \textbf{3.10} & \textbf{3.36} & \textbf{3.43} \\
\bottomrule
\end{tabular}
\caption{\textbf{Ablation on the Rewriter.} CA, CC, and RA denote Conversational Ability, Character Consistency, and Role-playing Attractiveness, respectively.}
\label{tab:ablation_rewriter}
\vspace{-0.2cm}
\end{table}

We conduct ablation studies on CharacterEval with Qwen2.5-7B-Instruct as the base model. To disentangle the contribution of profile and context edits, we compare three variants in Table~\ref{tab:ablation_rewriter}: rewriting only the profile, rewriting only the dialogue context, and rewriting both. The result shows that editing either part alone yields lower scores than editing both. The two rewriting targets stress different weaknesses of the Actor. Edited profiles test how well it adapts to perturbed character settings, while edited contexts test how it responds to unfamiliar dialogue states. Combining them gives the Rewriter more freedom to construct more diverse scenarios and improves the Actor's performance.

Besides, we compare three Rewriter settings in Table~\ref{tab:ablation_rewriter}. The overall score improves steadily from w/o Rewriter to frozen to co-trained, showing that rewriting itself brings a clear gain and that co-training brings a further one. Without RL updates, the frozen Rewriter cannot fully follow the Actor's evolving weaknesses, so part of its edits becomes less informative as training goes on. Co-training keeps the rewritten scenarios aligned with the current Actor and makes the adversarial loop between the two roles effective.

To test hyperparameter sensitivity, we vary the number of responses $N$ sampled per scenario when computing the Rewriter's reward (Equation~\ref{eq:rewriter_reward}) in Table~\ref{tab:ablation_n}. With $N{=}1$, the reward relies on a single response for each original and rewritten scenario, making the estimated weakness gap noisy. Increasing $N$ to $2$ recovers most of the gain and slightly surpasses $N{=}3$ on CC, while $N{=}3$ gives the best overall score. We therefore use $N{=}3$ as the default, as it provides a stable reward estimate with a moderate rollout cost.

\begin{table}[t]
\centering
\small
\setlength{\tabcolsep}{3.5pt}
\renewcommand{\arraystretch}{1.1}
\begin{tabular}{lcccc}
\toprule
\textbf{Variant} & \textbf{CA} ($\uparrow$) & \textbf{CC} ($\uparrow$) & \textbf{RA} ($\uparrow$) & \textbf{Overall} ($\uparrow$) \\
\midrule
$N{=}1$              & 3.79 & 3.07 & 3.30 & 3.39 \\
$N{=}2$              & 3.82 & \textbf{3.11} & 3.34 & 3.42 \\
$N{=}3$ (Ours)       & \textbf{3.83} & 3.10 & \textbf{3.36} & \textbf{3.43} \\
\bottomrule
\end{tabular}
\caption{\textbf{Ablation on $N$.} $N$ is the number of responses sampled per scenario in Equation~\eqref{eq:sample_n}.}
\label{tab:ablation_n}
\vspace{-0.2cm}
\end{table}

\subsection{Scenario Pool Diversity}
\label{sec:diversity}

We encode each scenario with text-embedding-3-large and quantify pool diversity with three metrics. Intra-Pool Distance (IPD) is the average pairwise cosine distance among all scenarios in the pool, reflecting overall spread. Embedding Drift (ED) is the average cosine distance between each rewritten scenario and its source, indicating how far the Rewriter pushes a scenario from its origin. New Clusters (NC) is the number of K-Means clusters that contain at least one rewritten scenario but no original scenario, measuring whether the pool expands into previously unoccupied semantic regions.

As shown in Table~\ref{tab:diversity}, all three metrics increase steadily across training epochs, indicating that the Rewriter keeps pushing scenarios into unexplored territory as the Actor improves. The supplementary material further analyzes whether these scenarios expose the current Actor's weaknesses and are later learned by the next Actor.

\section{Lanobe Benchmark}

Role-playing agents must handle a broad space of characters, situations, and speaking styles. The core problem is that a fixed training pool covers only a small part of this space, so the model can overfit to seen scenarios. Our experiments on English and Chinese benchmarks show that AdvRole helps by adding character and context pairs that expose model failures during training.

\begin{table}[t]
\centering
\small
\setlength{\tabcolsep}{3.5pt}
\renewcommand{\arraystretch}{1.1}
\begin{tabular}{lccc}
\toprule
\textbf{Pool} & \textbf{IPD} ($\uparrow$) & \textbf{ED} ($\uparrow$) & \textbf{NC} ($\uparrow$) \\
\midrule
$\mathcal{S}_0$ (Original)         & 0.564 & --    & --    \\
+ Epoch 1 Rewrite                  & 0.581 & 0.203 & 7/30  \\
+ Epoch 2 Rewrite                  & 0.594 & 0.231 & 11/30 \\
+ Epoch 3 Rewrite         & 0.602 & 0.254 & 14/30 \\
\midrule
+ Random Paraphrase                & 0.566 & 0.008 & 0/30  \\
\bottomrule
\end{tabular}
\caption{\textbf{Scenario pool diversity measured in embedding space.} NC is reported out of 30 total K-Means clusters fitted on the original pool. Random Paraphrase simply prompts GPT-4o to rephrase each scenario without any adversarial objective.}
\label{tab:diversity}
\vspace{-0.2cm}
\end{table}

This issue is amplified in multilingual settings, where character names, backgrounds, scenes, and dialogue styles are often language-specific. Existing role-playing benchmarks mainly cover English~\cite{coser_icml25} or Chinese~\cite{charactereval_acl24, wu-etal-2025-raiden}, so they cannot show whether improving the scenario pool also helps when the language and cultural setting change. To test this, we introduce \textbf{Lanobe}, a multilingual role-playing benchmark covering five languages: Simplified Chinese, Traditional Chinese, Japanese, Korean, and Thai. For each language, we provide 1,000 training and 500 test scenarios.

We collect character descriptions and plot summaries from publicly available sources and process them through a multi-stage LLM pipeline powered by Kimi-K2.6.\footnote{The released benchmark contains only LLM-generated text. No verbatim excerpts from copyrighted material are included, and the benchmark will be released under a non-commercial license for research purposes only.} The full construction pipeline is described in the supplementary material.

For evaluation, we use Kimi-K2.6 as the LLM judge to score each model response along four dimensions on a 0--10 scale: Character Consistency (CC), Situational Coherence (SC), Linguistic Quality (LQ), and Interaction Proactivity (IP). The final score is the arithmetic mean of the four dimensions.

\begin{table}[t]
\centering
\small
\setlength{\tabcolsep}{3.5pt}
\renewcommand{\arraystretch}{1.1}
\begin{tabular}{lccccc}
\toprule
\textbf{Qwen2.5-7B-Instruct} & \textbf{CC} & \textbf{SC} & \textbf{LQ} & \textbf{IP} & \textbf{Avg.} \\
\midrule
\multicolumn{6}{l}{\textit{Simplified Chinese}} \\
\quad Vanilla                          & 5.99 & 5.70 & 7.59 & 5.04 & 6.08 \\
\quad CPO                              & 6.19 & 5.85 & 7.65 & 5.29 & 6.24 \\
\quad R4                               & 6.11 & 5.65 & 7.62 & 4.96 & 6.09 \\
\quad VeriRole                         & 6.34 & 5.78 & 7.54 & 5.19 & 6.21 \\
\quad \textbf{AdvRole (Ours)}             & \textbf{6.41} & \textbf{6.05} & \textbf{7.71} & \textbf{5.49} & \textbf{6.42} \\
\midrule
\multicolumn{6}{l}{\textit{Traditional Chinese}} \\
\quad Vanilla                          & 5.09 & 4.98 & 6.01 & 4.78 & 5.21 \\
\quad CPO                              & 5.29 & 5.13 & 6.07 & 5.03 & 5.38 \\
\quad R4                               & 5.21 & 4.93 & 6.04 & 4.70 & 5.22 \\
\quad VeriRole                         & 5.44 & 5.06 & 5.96 & 4.93 & 5.35 \\
\quad \textbf{AdvRole (Ours)}             & \textbf{5.51} & \textbf{5.33} & \textbf{6.13} & \textbf{5.23} & \textbf{5.55} \\
\midrule
\multicolumn{6}{l}{\textit{Japanese}} \\
\quad Vanilla                          & 5.32 & 4.98 & 6.95 & 4.52 & 5.44 \\
\quad CPO                              & 5.52 & 5.13 & 7.01 & 4.77 & 5.61 \\
\quad R4                               & 5.44 & 4.93 & 6.98 & 4.44 & 5.45 \\
\quad VeriRole                         & \textbf{5.67} & 5.06 & 6.90 & 4.67 & 5.58 \\
\quad \textbf{AdvRole (Ours)}             & \textbf{5.67} & \textbf{5.26} & \textbf{7.03} & \textbf{4.90} & \textbf{5.71} \\
\midrule
\multicolumn{6}{l}{\textit{Korean}} \\
\quad Vanilla                          & 4.70 & 4.45 & 6.42 & 4.00 & 4.89 \\
\quad CPO                              & 4.90 & 4.60 & 6.48 & 4.25 & 5.06 \\
\quad R4                               & 4.82 & 4.40 & 6.45 & 3.92 & 4.90 \\
\quad VeriRole                         & \textbf{5.05} & 4.53 & 6.37 & 4.15 & 5.03 \\
\quad \textbf{AdvRole (Ours)}             & 5.00 & \textbf{4.70} & \textbf{6.49} & \textbf{4.35} & \textbf{5.13} \\
\midrule
\multicolumn{6}{l}{\textit{Thai}} \\
\quad Vanilla                          & 5.49 & 5.04 & 6.51 & 4.97 & 5.50 \\
\quad CPO                              & 5.69 & 5.19 & \textbf{6.57} & 5.22 & 5.67 \\
\quad R4                               & 5.61 & 4.99 & 6.54 & 4.89 & 5.51 \\
\quad VeriRole                         & \textbf{5.84} & 5.12 & 6.46 & 5.12 & 5.64 \\
\quad \textbf{AdvRole (Ours)}             & 5.74 & \textbf{5.24} & 6.56 & \textbf{5.25} & \textbf{5.70} \\
\bottomrule
\end{tabular}
\caption{\textbf{Performance on the Lanobe benchmark.} CC, SC, LQ, and IP denote Character Consistency, Situational Coherence, Linguistic Quality, and Interaction Proactivity, respectively. All dimensions are scored by Kimi-K2.6 on a 0--10 scale.}
\label{tab:lanobe}
\vspace{-0.2cm}
\end{table}

We train all methods on the Lanobe training set using the same setting described in the Implementation Details section. As shown in Table~\ref{tab:lanobe}, AdvRole achieves the highest average score across all five languages, confirming that the adversarial scenario-rewriting strategy generalizes beyond English-centric benchmarks. Notably, VeriRole obtains the highest CC score on Korean and Thai, consistent with its character-consistency-focused reward design observed in CharacterEval. However, AdvRole still achieves the best overall score in every language by improving more broadly across all four dimensions.

These results also show why multilingual evaluation is needed. AdvRole brings clear gains in Chinese, while the gains on Korean and Thai are smaller. This suggests that improving the scenario pool is not the only factor in multilingual role-playing. The base model's language ability also matters. When the base model already has stronger language knowledge, RL can better turn targeted training scenarios into role-playing gains. When the target language is weaker in the base model, the same training signal has less room to work.

\section{Conclusion}

In this paper, we address the issue that role-playing agents are limited by a fixed pool of training scenarios. We propose AdvRole, an adversarial RL framework in which a Rewriter edits character profiles and dialogue contexts into scenarios that expose the Actor's weaknesses and is rewarded by the Actor's performance gap between the original and rewritten ones. Experiments on three role-playing benchmarks, including a new multilingual one we release, show that AdvRole outperforms baselines and offers a new perspective on training role-playing agents by evolving the training data.

\bibliography{aaai2027}

\end{document}